\documentclass{article}

\usepackage{arxiv}

\usepackage{tabularx}
\usepackage{caption}
\usepackage{subcaption}
\usepackage{multicol}
\usepackage{multirow}
\usepackage{silence}
\usepackage[linesnumbered,ruled]{algorithm2e}

\usepackage[utf8]{inputenc} 
\usepackage[T1]{fontenc}    
\usepackage{hyperref}       
\usepackage{url}            
\usepackage{booktabs}       
\usepackage{amsfonts}       
\usepackage{nicefrac}       
\usepackage{microtype}      
\usepackage{lipsum}
\usepackage{graphicx}
\graphicspath{ {./images/} }

\title{The Severity Prediction of The Binary And Multi-Class Cardiovascular Disease - A Machine Learning-Based Fusion Approach}

\author{
Hafsa Binte Kibria  \\
   Department of Electrical \& Computer Engineering\\
   Rajshahi University of Engineering \& Technology, \\
   Rajshahi-6204,Bangladesh \\
   \texttt{hafsabintekibria@gmail.com} \\
   \And
Abdul Matin \\
   Department of Electrical \& Computer Engineering\\
   Rajshahi University of Engineering \& Technology, \\
   Rajshahi-6204,Bangladesh \\
   \texttt{ammuaj.cseruet@gmail.com} \\
}

\begin{document}
\maketitle
\begin{abstract}
In today's world, a massive amount of data is available in almost every sector. This data has become an asset as we can use this enormous amount of data to find information. Mainly health care industry contains many data consisting of patient and disease-related information. By using the machine learning technique, we can look for hidden data patterns to predict various diseases. Recently CVDs, or cardiovascular disease, have become a leading cause of death around the world. The number of death due to  CVDs is frightening. That is why many researchers are trying their best to design a predictive model that can save many lives using the data mining model. In this research, some fusion models have been constructed to diagnose CVDs along with its severity. Machine learning(ML) algorithms like artificial neural network, SVM, logistic regression, decision tree, random forest, and AdaBoost have been applied to the heart disease dataset to predict disease. Randomoversampler was implemented because of the class imbalance in multiclass classification. To improve the performance of classification, a  weighted score fusion approach was taken. At first, the models were trained. After training, two algorithms' decision was combined using a weighted sum rule. A total of three fusion models have been developed from the six ML algorithms. The results were promising in the performance parameter. The proposed approach has been experimented with different test training ratios for binary and multiclass classification problems, and for both of them, the fusion models performed well. The highest accuracy for multiclass classification was found as 75\%, and it was 95\% for binary. The code can be found in : \href{https://github.com/hafsa-kibria/Weighted_score_fusion_model_heart_disease_prediction} {github/hafsakibria/Weightedscorefusion} 

\end{abstract}


\section{Introduction}
Cardiovascular diseases (CVD) are caused by the heart and blood vessels \cite{wong2014epidemiological}. It is one of the significant causes of death around the world. Annually, it kills more people than any other cause. In 2016, the death rate was around 17.9 million, which represents 31\% of worldwide mortality. Among them, 85\% are because of stroke and heart attack. Most cardiovascular disorders can be avoided by controlling some approaches like unhealthy diet, imbalanced lifestyle, physical inactivity, and excessive alcohol usage. According to the estimation of WHO, almost 23.6 million deaths will occur because of CVDs, mainly from stroke and heart disease, by 2030. This estimation shows that cardiovascular disease remains the single major cause of death until now. So people with a high risk of cardiovascular diseases need to be detected. That is why diagnosing the disease correctly is necessary and proper treatment can also reduce the risk of further severity of the disease. A clear understanding of the risk is required to improve the diagnosis. In the traditional approach, diseases are identified by analyzing patients' symptoms and medical history, for example, ECG tests, blood sugar levels, blood pressure, cholesterol. This process is time-consuming and expensive. With the aid of machine learning, it becomes simple. This process saves much time and therefore improves the efficiency of the diagnosis. This is one of the reasons we have developed some fusion models to classify heart disease efficiently. Here heart disease has been classified using some machine learning-based fusion models.
        
The world data is growing day by day, and hospitals are slowly adopting big data systems \cite{gans2005medical}. Applying data analysis in the medical sector is giving excellent benefits. It improves the result and reduces cost. Effective implementation of machine learning improves physicians' work and increases the productivity of the healthcare service. In diagnosing clinical data, significant improvement has been observed using machine learning. Several diseases have been predicted using machine learning like diabetes, heart disease, breast cancer \cite{ amrane2018breast, palaniappan2008intelligent}. Machine Learning is closely linked to both statistics and decision-making. They can be used for several purposes, such as forecasting the amount of product sold, forecasting covid cases in the upcoming month, the probability of rainfall occurring in a particular area, selling airline tickets, etc. \cite{kibria2021forecasting, cramer2017extensive}. As the medical sector has a large dataset, these existing data will help the researchers diagnose the disease early by systematically analyzing data \cite{harimoorthy2020multi}. We have analyzed the binary and multi-classification of heart disease in this research using existing data.

The primary purpose of this study is to classify patients with heart disease using medical records. The five-class classification model can also predict the severity stage of patients with heart disease. We have used six different machine learning algorithms to classify cardiovascular disease in our proposed approach. First, the data was pre-processed. Then the individual algorithms were trained with the same dataset. The weighted sum rule was applied to the decision score provided by the trained algorithms. The fusion models used the new score to classify disease. We have developed one fusion model from two individual algorithms. So in this work, three weighted score fusion models were generated, which provided an improving performance compared to the previous separate machine learning algorithms. A machine learning-based predictive system can reduce physicians' pressure, and a fusion approach will help diagnose disease more efficiently.

\section{Related Work}
Nowadays, many researchers around the world are focusing on machine learning algorithms in the health field to forecast different diseases. And the use of machine learning in the medical sector has given a notable change in the performance of treatment. This section discusses the various diseases such as diabetes, cardiovascular, kidney, which have been diagnosed using machine learning algorithms.


Comparative analysis of various diseases has been done using machine learning algorithms \cite{dinh2019data,singh2020heart,ali2021heart, shah2020heart, diwakar2021latest, kibria2021comparative}. In \cite{dinh2019data}, a comparative analysis was developed using machine learning algorithms. Both cardiovascular and diabetes disease datasets were used for classification. Different machine learning algorithms like XGBoost, random forest, and weighted ensemble models were used to predict disease. The essential features that contribute most to the dataset were identified. At last, the performance parameters had been shown to observe the results. Ensemble models showed slightly better accuracy than other models. The researchers also proposed to apply the model in a real-world scenario to check the risk of the disease occurring. They used ten-fold cross-validation, and its accuracy was slightly low. So to increase the accuracy, we have used a fusion of two algorithms rather than a single algorithm.

Two data mining algorithms have been used \cite{ayatollahi2019predicting} for the diagnosis of cardiovascular disease. Among them, the SVM showed a strong efficiency. The specificity of SVM and ANN was 74.42\% and 73.64\%, respectively. Here performance for both algorithms was poor. Using the same dataset, we have been able to get a higher accuracy for every algorithm. And after fusion, the accuracy further improved in our work. Decision-making architectures was introduced in \cite{arabasadi2017computer, gavhane2018prediction, tarawneh2019hybrid, iftikhar2017evolution}. In \cite{arabasadi2017computer} a computer-based decision-making architecture was applied with a genetic algorithm. Performance improved significantly after combining the genetic algorithm with an artificial neural network.

Multiple diseases such as cardiovascular, chronic kidney disease (CKD), and diabetes were identified  in \cite{harimoorthy2020multi, xiao2019comparison, hasan2020diabetes, qiao2022rlds}. Support vector, decision tree, and random forest algorithms were used for classification with a standardized decision support model in \cite{harimoorthy2020multi}. A Chi-square method was employed to select the best features. They implemented SVM (linear, polynomial, and radial) procedures using extracted features. The performance was evaluated with the help of accuracy, specificity, miss classification rate classification parameters, and other parameters. Improved SVM-Radial gave the best accuracy of all the algorithms.  We got a better accuracy using weighted score fusion.

A modern gender-based approach was built in \cite{hogo2020proposed}. Various classifiers had been applied, and multi-layer perceptron achieved better results than any other data mining techniques. In another study, \cite{mir2018diabetes}, four data mining algorithms were applied to the diabetes dataset using the WEKA tool. The aim was to evaluate the performance of all algorithms and find out the best one for prediction. The four classifiers were compared, and among them, SVM gave the highest f1-score of 78.2\%. The overall performance of SVM was found better than Random Forest, Naive Bayes, and Simple Cart. Survival of heart patients was predicted in \cite{mehedi2021survival}.

In \cite{thomas2016human}, the data mining technique has been used to diagnose heart disease. KNN and ID3 algorithms were applied. Different accuracy had been observed with the change in the number of attributes. A comparative analysis has been shown in \cite{singh2016heart} with different machine learning algorithms. Different validation techniques were applied to evaluate the performance. Support vector machine, random forest, decision tree, linear and logistic regression were used for prediction. The highest accuracy achieved by the proposed model was 85.81\%  using ten-fold cross-validation by random forest.

In \cite{birjais2019prediction}, KNN imputation was used to handle missing values, and gradient boosting, logistic regression, and naive Bayes had been used to predict diabetes disease. Accuracy, specificity, and sensitivity were measured for performance measurement and validation. Gradient boosting gave the best result for prediction. A computer-based support tool was introduced for diabetics prediction in this paper \cite{tafa2015intelligent}. Here two data mining algorithms, SVM and Naive Bayes were applied in Matlab. The individual decision was taken from both algorithms, and then the joint outcomes from both algorithms were taken for a new hybrid model. The goal of the hybrid model was to make the system more reliable. It minimized the false-negative response of the system. In a recent study \cite{kibria2020efficient},\cite{kibriaa2020comparative} decision level fusion model has been introduced for the classification of heart disease. They have merged two algorithms for classification and obtained a better result than a single algorithm. We used weighted score fusion instead of decision level fusion to get a more improved accuracy.
 
These studies show that the researchers have used various machine learning algorithms to classify the disease. A limitation of these works is that only binary classification has been done for heart disease classification. It is also essential to determine the result of multi-classification to know what stage a heart disease patient is currently present. So we have classified both two and five-class classifications. And, there is a scope for further improvement in accuracy in the discussed methods. That is why scientists are developing fusion models to classify the disease more accurately. In our work, three weighted fusion models have been created to observe if they can give better results than the separate models that have been used in the fusion. The weighted sum rule was implemented to combine the decision of two individual algorithms \cite{matin2017weighted}. And in our output, we have observed a good improvement in almost every parameter of the fusion models. For two and five-class classifications, we built weighted fusion models and got an improving performance.

\section{Materials and Methodology}

\subsection{Proposed Approach}
We conducted quantitative research in this case. Since our goal is to predict disease, we used a quantitative method. We used secondary data in our research and solved a real-world problem. Other researchers have also used this data. In our work, the focus is on determining the seriousness of patients’ heart disease along with the presence and absence of it. Two classification models (two and five classes) were developed. First, a multi-classification (five class) system was built to measure the disease’s severity. And the binary classification only classified the patients with or without heart disease.
    
 \begin{figure}[ht]
\centerline{\includegraphics[width=.9\columnwidth]{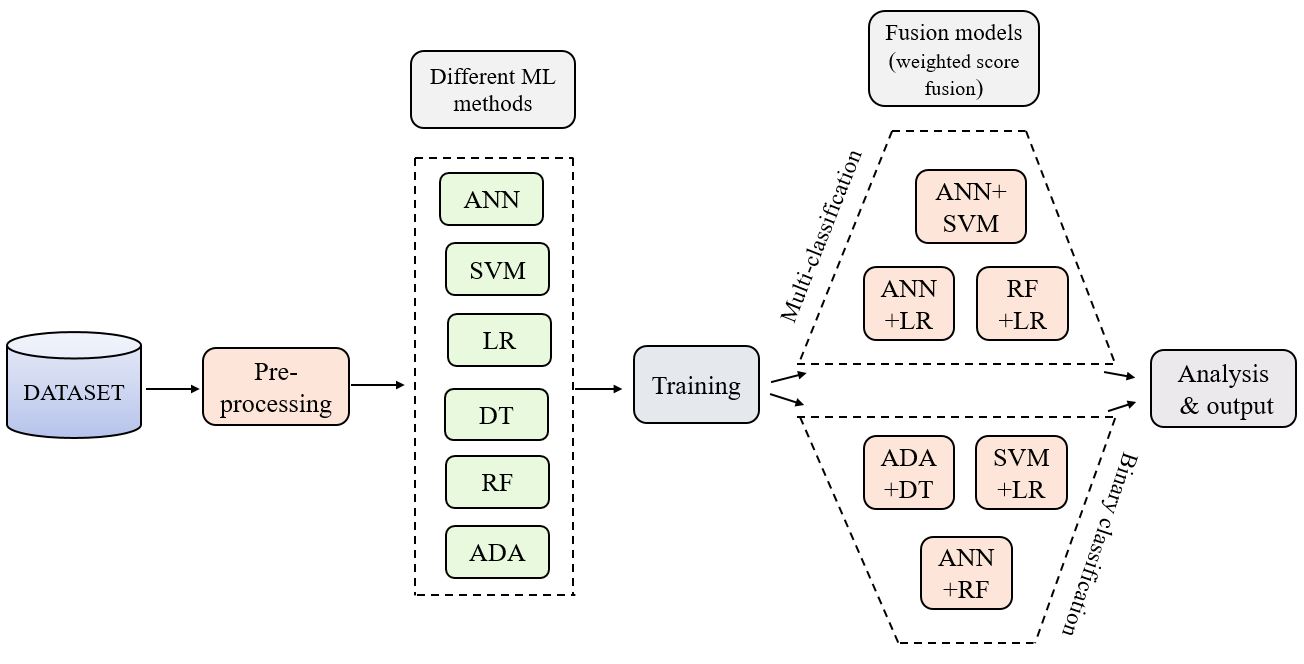}}
\caption{Proposed architecture}
 \label{pa}
\end{figure}

Six machine learning algorithms were used, and three fused models were developed from the six ML models. The algorithms were selected based on their performance. It was noticed that algorithms with accuracy near each other tend to give a better result after fusion. In most cases, this strategy was followed. We did not follow the above-mentioned strategy for only two cases as it did not give a satisfactory outcome, so we randomly checked which combination gave better results and went for it. The two cases mentioned above were for fusion model-1 in multi-classification (80:20 split) \& fusion model-1 for binary classification (70:30 split).

First, the data have been pre-processed, which means it was cleaned, missing values were replaced by the most frequent strategy, and feature scaling was implemented. After cleaning, the data were divided into test and train sets. Then, we trained the dataset with six individual algorithms. As the dataset was imbalanced for multi-class classification, oversampling technique was used to balance the dataset. Now for fusion, the decision from two algorithms was added using the weighted sum rule. Before merging, standardization was unnecessary as every individual model’s output was in the same score range.

The complete proposed model has been shown in Figure \ref{pa}. The arrangement of fusion models for binary and multi-classification was slightly different.

\begin{figure}[htbh]
\centerline{\includegraphics[width=.9\columnwidth]{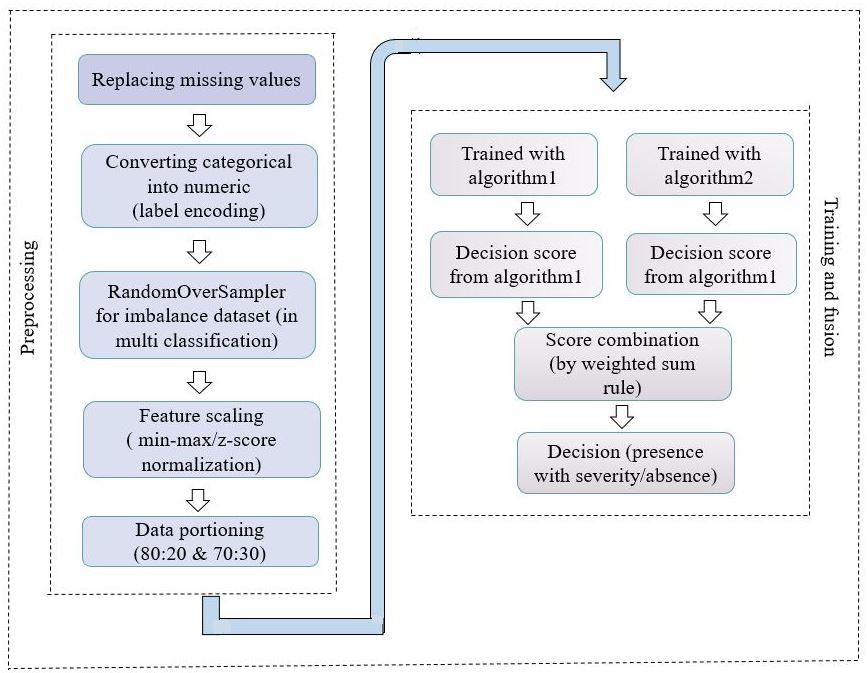}}
\caption{Overall flow chart of weighted score level fusion model}
\label{over}
\end{figure}

Figure \ref{over} shows the overall flow chart for heart disease classification. In the first step, data were pre-processed, and the next step was for training and fusion. This is the generalized structure for weighted score fusion, and this approach was used for all the fusion models using different algorithms. The steps mentioned in figure \ref{over} are listed below in detail.

\subsubsection{Data description}
\begin{figure}[htbh]
\centerline{\includegraphics[width=1\columnwidth]{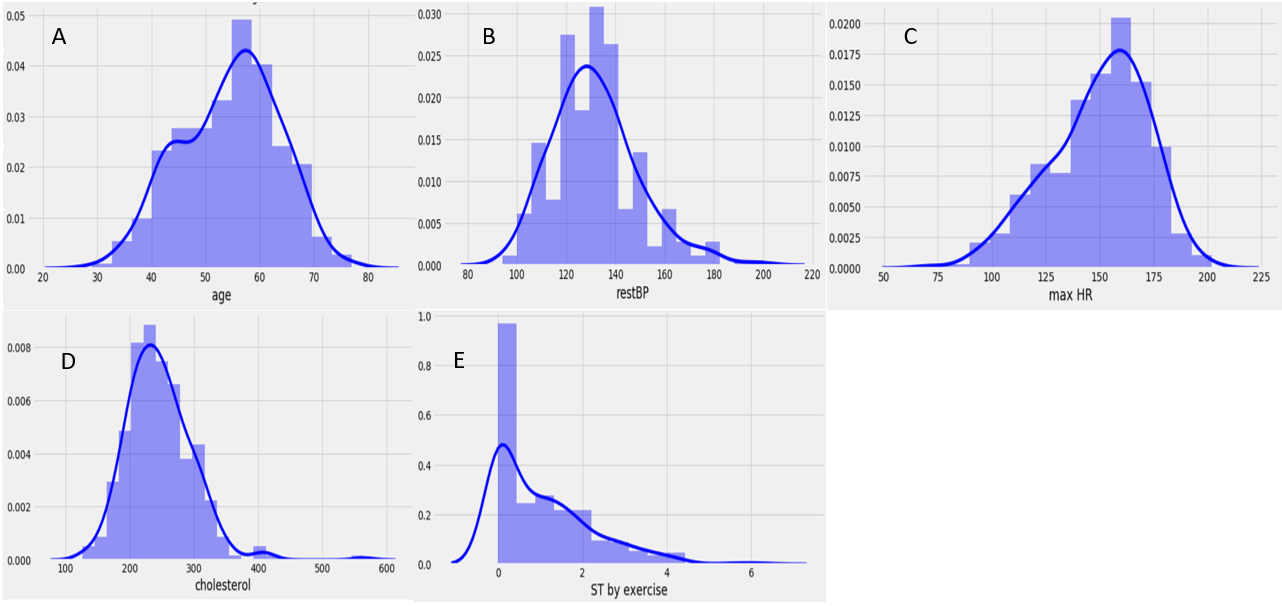}}
\caption{Data distribution for continuous attributes A) Distribution of age B) Resting blood pressure C) Maximum heart rate D) Cholesterol E) Oldpeak}
\label{dd}
\end{figure}

Dataset of the heart disease was taken from the archive of Cleveland UCI \cite{Dua}. It has 303 cases, 76 attributes, and other missing properties. The most frequent values have been replaced with the missing values in the dataset. Researchers typically use 13 of the 76 attributes. The output contains five labels with no cardiac disease(0) and other values(1-4) reflecting cardiac severity. Categorical values in the dataset were converted into numeric using label encoder. Table \ref{data} below provides a detailed description of each attribute’s mean value and standard deviation. It also shows the data distribution of the categorical value in the dataset. The data for training and testing was split into 70:30 and 80:20; for both ratios, output parameters were calculated. Figure \ref{dd} shows the data distribution of the continuous value from the dataset.

\begin{figure}[htbh]
\centering
\begin{subfigure}{.35\textwidth}
  \centering
  \includegraphics[width=.95\linewidth]{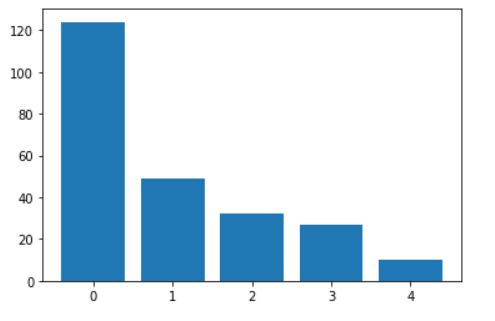}
  \caption{Before oversampling}
  \label{1}
\end{subfigure}%
\begin{subfigure}{.35\textwidth}
  \centering
  \includegraphics[width=.95\linewidth]{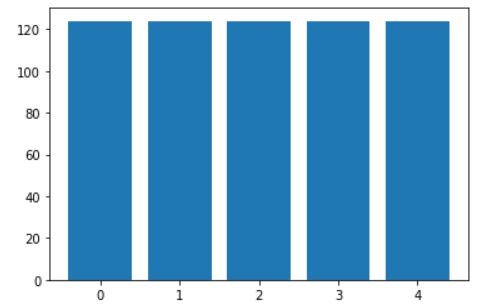}
  \caption{After oversampling}
  \label{2}
\end{subfigure}
\caption{Distribution of target class of training set before and after oversampling}
\label{resample}
\end{figure}

The output data was balanced for binary classification but unbalanced for multi-class classification. There are some techniques for the imbalanced dataset like oversampling, undersampling, etc. The random oversampling technique has been used for our dataset in multi-class classification. 
In undersampling, it removes potential valuable data. That is why we have decided to do oversampling. A random set of copies of minority class examples is added to the data in oversampling. That is how it makes the balance between the minority and majority classes. Figure \ref{1} represents the distribution of our training set before making the dataset balanced, and figure \ref{2} is the representation after making them balanced. We first resampled the training dataset using oversampling, and then our model was trained on that dataset. After training, it was used for the prediction of the test dataset. The RandomOverSampler algorithm was applied only on the training dataset. Otherwise, it will create new data in the test set.

\begin{table*}[t!]
\setlength{\tabcolsep}{3pt}
\centering
\caption{Description of Features }
\begin{tabular}{|c|c|l|c|c|c|}
\hline
\multirow{2}{*}{\textbf{S. No.}} & \multirow{2}{*}{\textbf{Attribute}} & \multicolumn{1}{c|}{\multirow{2}{*}{\textbf{Description}}}                                                                                                                                                                            & \multirow{2}{*}{\textbf{\begin{tabular}[c]{@{}c@{}}Data \\ distribution\end{tabular}}} & \multirow{2}{*}{\textbf{Mean}} & \multirow{2}{*}{\textbf{Std}} \\
                                 &                                     & \multicolumn{1}{c|}{}                                                                                                                                                                                                                 &                                                                                        &                                &                               \\ \hline
1                                & Age                                 & Age in years(29-77)                                                                                                                                                                                                                   & continuous                                                                             & 54.4                           & 9.07                          \\ \hline
2                                & Sex                                 & \begin{tabular}[c]{@{}l@{}}Gender of patient(1 is represented as male and 0\\ as female)\end{tabular}                                                                                                                                 & \begin{tabular}[c]{@{}c@{}}0=32.01\%\\ 1=67.98\%\end{tabular}                          & 0.68                           & 0.47                          \\ \hline
3                                & Cpt                                 & \begin{tabular}[c]{@{}l@{}}Chest pain type is   classified into 4 values.\\ 1.typical   angina 2.atypical angina\\ 3.non-angina pain 4. asymptomatic\end{tabular}                                                                     & \begin{tabular}[c]{@{}c@{}}1=47.52\%\\ 2=16.50\%\\ 3=28.38\%\\ 4=7.59\%\end{tabular}   & 0.97                           & 1.03                          \\ \hline
4                                & Thstbps                             & \begin{tabular}[c]{@{}l@{}}Resting blood pressure (in   mm Hg following\\ hospital admission)with range of (94-200)\end{tabular}                                                                                                      & continuous                                                                             & 132                            & 17.5                          \\ \hline
5                                & S\_chol                             & Serum cholesterol in   mm/dl(126-564)                                                                                                                                                                                                 & continuous                                                                             & 246                            & 51.7                          \\ \hline
6                                & Restelect                           & \begin{tabular}[c]{@{}l@{}}Results of resting electrocardiography or   ECG is \\ divided into 3 values.0 is represented as normal, 1\\ as abnormal ST T   wave and 2 as the response\\ of left ventricular   hypertrophy\end{tabular} & \begin{tabular}[c]{@{}c@{}}0=1.32\%\\ 2=48.84\%\\ 3=49.83\%\end{tabular}               & 0.53                           & 0.52                          \\ \hline
7                                & FBS                                 & \begin{tabular}[c]{@{}l@{}}if level of fasting blood   sugar is greater equal \\ 120, represented as 1 for true, 0 for false\end{tabular}                                                                                             & \begin{tabular}[c]{@{}c@{}}0=85.47\%\\ 1=15.18\%\end{tabular}                          & 0.15                           & 0.36                          \\ \hline
8                                & thlach                              & \begin{tabular}[c]{@{}l@{}}the maximum heart rate that   is achieved, in the\\ range between 71 to 202.\end{tabular}                                                                                                                  & continuous                                                                             & 150                            & 22.9                          \\ \hline
9                                & Exng                                & \begin{tabular}[c]{@{}l@{}}Angina caused by exercise(0 depicted as no and   \\ 1 as yes)\end{tabular}                                                                                                                                 & \begin{tabular}[c]{@{}c@{}}0=67.32\%\\ 1=32.67\%\end{tabular}                          & 0.33                           & 0.47                          \\ \hline
10                               & Oldpeak                             & \begin{tabular}[c]{@{}l@{}}ST depression caused by exercise relative to rest, \\ continuous value in the range between 0 to 6.2\end{tabular}                                                                                          & continuous                                                                             & 1.04                           & 1.16                          \\ \hline
11                               & Slp                                 & Slope of the peak exercise   ST segment                                                                                                                                                                                               & \begin{tabular}[c]{@{}c@{}}0=6.93\%\\ 1=46.20\%\\ 2=46.86\%\end{tabular}               & 1.4                            & 0.62                          \\ \hline
12                               & Ca                                  & \begin{tabular}[c]{@{}l@{}}Number of major vessels   that are colored by \\ floursopy from 0 to 3\end{tabular}                                                                                                                        & \begin{tabular}[c]{@{}c@{}}0=59.40\%\\ 1=21.45\%\\ 2=12.54\%\\ 3=6.60\%\end{tabular}   & 0.73                           & 1.02                          \\ \hline
13                               & Thal                                & \begin{tabular}[c]{@{}l@{}}Blood disorder represented with three distinct \\ numbers.3 as normal,6 as fixed and 7 as\\  reversible defect\end{tabular}                                                                                & \begin{tabular}[c]{@{}c@{}}3=5.94\%\\ 6=55.44\%\\ 7=38.61\%\end{tabular}               & 2.31                           & 0.61                          \\ \hline
14                               & Output                              & \begin{tabular}[c]{@{}l@{}}Classification of heart disease is shown into five \\ values,0 as absence and 1-4   represents the degree \\ of severity\end{tabular}                                                                      & \begin{tabular}[c]{@{}c@{}}0=54.12\%\\ 1-4=45.87\%\end{tabular}                        & 0.54                           & 0.5                           \\ \hline
\end{tabular}
\label{data}
\end{table*}

\subsubsection{Data Transformation}
Data needs to be in an appropriate format for the data mining process. That is why data transformation or standardization is necessary. It reduces the redundancy of data. It is the process of changing data structure, configuration, and values of data. Data standardization is a method by which one or more attributes are rescaled to have a mean value of 0 and a standard deviation of 1. In SVM or logistic regression, if one function has a higher value than others, it will dominate other features. So to have the same influence on every attribute, data transformation is essential. It makes data better organized and also improves data quality. Both standardization (z-score) and min-max normalization have been used in this work to transform data for different algorithms. The standard score for a sample x is determined as:

    \begin{equation}
    q=(x-r)/d
    \end{equation}
    where r is the mean of training samples, d is the standard deviation. 
    And the equation for min-max scaler is:
    \begin{equation}
    h'=\frac{h-min(h)}{max(h)-min(h)}
    \end{equation}    
    Where h is an original value, and h' is the normalized value.


\subsubsection{Weighted score fusion}
After training all six algorithms, this step was implemented at the decision level. We got a decision probability $(Di)$  for each algorithm to predict the test data. From this decision score, the prediction was made for the test data.
Different weights have been assigned to each algorithm’s decision probability so that each algorithms’ effect varies in the fusion model. If one of the algorithms gave a higher rate for the right decision, we assigned a bigger weight to that and a comparatively smaller weight to the other algorithm. So if one of the algorithms has a weight of .75, then the other will have (1-.75)=.25. We used a loop to check which weights provided the best accuracy for the fusion model and selected them. The sum of the weights used in the fusion model should be 1 for scaling. We selected the weights that gave the best result for the fusion model. The equation for the weighted sum is as follows:

    \begin{equation}
      D_f =\displaystyle\sum^n_{i=1} W_i*D_i
      \label{eq3}
   \end{equation}

Where n is the number of the algorithm used for fusion, we have used two algorithms for every fusion model, so n=2. $D_f$ is the weighted sum, which is the new decision score of the weighted score fusion model. Based on this score, the final decision was given. $W_i$ represents the weight that has been assigned to an algorithm with the decision score, and $D_i$ is the decision probability score for any individual algorithm.

The process of building a fusion model is illustrated below. Two models were used to create a fused model. A weighted score level fusion model was developed by merging those single algorithms, and it worked better than the individual algorithm. The algorithm for weighted level fusion is given below:

\begin{algorithm}
    \SetKwInOut{Input}{Input}
    \SetKwInOut{Output}{Output}

    \Input{ Two float value $D_i$,$W_i$, one int $n$. $D_i$ is the individual Algorithm's decision score after training,$W_i$ is the corresponding weights of the decision scores, $n$ is the number of separate algorithms in fusion}  
    \Output{$D_f$,new decision score of weighted score fusion model}
     $ W_1=1 $  \\
  \For{$i\gets0$ \KwTo 19 } {   
        $D_f=0$
        \\
 $ W_1=W_1-0.05$,
 $ W_2=1-W_1 $   \tcp*{ $ \displaystyle\sum^n_{i=1} W_i=1 $ } 
         \For{$i\gets0$ \KwTo $n$ }{
       $ D_f =\displaystyle\sum W_i*D_i$
    }
    }
Select the weights ($W_1,W_2$) that gave the highest decision score.\\
 $ D_f =\displaystyle\sum W_i*D_i$  \tcp*{Final fused decision score using selected weights}
    
    \caption{Algorithm for weighted score fusion}
\end{algorithm}

After selecting weights, the weighted sum rule was applied at the last step in the algorithm. A weight was assigned to each individual model’s decision score in a weighted sum. To select the weights, we used a loop for using various values of weights shown in table \ref{weicom}, and the weights that gave the highest decision score were selected. For example, there were two models in our approach. The individual result was fused, but the outcome of the two algorithms was not taken equally. Rather than weights were assigned that decided the effect of any algorithm in the fusion model.

Then from the new decision score, the result of our fusion model was predicted. The code is given \href{https://github.com/hafsa-kibria/Weighted_score_fusion_model_heart_disease_prediction}{here}. The steps for the fusion is given below in detail:

\begin{enumerate}
\item	The output of any algorithm was determined from the prediction scores. It ranges between 0 to 1. At first, the prediction scores ($D1$ \& $D2$) of the two algorithms were taken.
\item	Then using the selected weights($W1$ \& $W2$), the final decision score ($Df$) was obtained using equation \ref{eq3}. From this score, we got our output value. If the final decision score was greater than .5, the output was zero else one for binary classification.
\item	A for loop was used to select the weights from the different combinations shown in the table \ref{weicom}. The final decision score ($Df$) of the test set was calculated using the different combinations of weights ($W1$ \& $W2$) and using the two algorithms’ given decision probability scores ($D1$ \& $D2$). The weights that showed the highest decision score ($Df$) were selected for fusion among the combinations.
\item	After selecting the weights, they were assigned to the decision scores of the individual models like in the equation \ref{eq3}, and the fusion decision score was calculated for both binary and multi-class classification.
\item	The output was calculated from this final decision score ($Df$). There was a decision score for every test data by which the prediction was made. 
\begin{enumerate}
    \item The new decision score ($Df$) of the fusion model contained two different float numbers for every test data in binary class, and the sum of the values was one. The index of the higher value indicates the target class of that test data. The prediction was made by extracting the index number of the greater value.
    
    \item	Each algorithm’s prediction score contained five float numbers representing the five different classes for multi-class classification. The final decision score had five different values in an array for every test data. The index containing the maximum value was the target class. We extracted the index number from the array to know the predicted output.
\end{enumerate}

\item At last, the prediction was made for the test data from the final decision score ($Df$).
\end{enumerate}

We used the heart disease dataset in our proposed framework for both binary and multi-classification problems. In multi-classification, data were split into multiple groups where there were two classes in binary classification. Performance parameters were measured to see how well the models performed.

\begin{table}[h]
\centering
\caption{Combinations of weights used in fused algorithms}
\begin{tabular}{|c|c|c|c|}
\hline
\textbf{\begin{tabular}[c]{@{}c@{}}Number \\ of loop\end{tabular}} & \textbf{W1} & \textbf{W2} & \textbf{\begin{tabular}[c]{@{}c@{}}Sum of\\  weights\end{tabular}} \\ \hline
1                                                                  & 0.95        & 0.05        & 1                                                                  \\ \hline
2                                                                  & 0.9         & 0.1         & 1                                                                  \\ \hline
3                                                                  & 0.85        & 0.15        & 1                                                                  \\ \hline
4                                                                  & 0.8         & 0.2         & 1                                                                  \\ \hline
5                                                                  & 0.75        & 0.25        & 1                                                                  \\ \hline
6                                                                  & 0.7         & 0.3         & 1                                                                  \\ \hline
7                                                                  & 0.65        & 0.35        & 1                                                                  \\ \hline
8                                                                  & 0.6         & 0.4         & 1                                                                  \\ \hline
9                                                                  & 0.55        & 0.45        & 1                                                                  \\ \hline
10                                                                 & 0.5         & 0.5         & 1                                                                  \\ \hline
11                                                                 & 0.45        & 0.55        & 1                                                                  \\ \hline
12                                                                 & 0.4         & 0.6         & 1                                                                  \\ \hline
13                                                                 & 0.35        & 0.65        & 1                                                                  \\ \hline
14                                                                 & 0.3         & 0.7         & 1                                                                  \\ \hline
15                                                                 & 0.25        & 0.75        & 1                                                                  \\ \hline
16                                                                 & 0.2         & 0.8         & 1                                                                  \\ \hline
17                                                                 & 0.15        & 0.85        & 1                                                                  \\ \hline
18                                                                 & 0.1         & 0.9         & 1                                                                  \\ \hline
19                                                                 & 0.05        & 0.95        & 1                                                                  \\ \hline
\end{tabular}
\label{weicom}
\end{table}


\subsection{Modeling}

\subsubsection{ Logistic regression(LR)}
Logistic regression is a specific form of GLM, which is frequently abbreviated as a Generalized Linear Model. It is like linear regression, but it predicts true or false. Instead of adjusting a line to the data, LR provides a s-shaped logistic function. The output probability for any given problem is found from the curve, and it is generally used for binary classification. Logistic regression can work with both continuous and discrete data. In linear regression, the line is fit by least squares, whereas logistic regression uses maximum likelihood. The likelihood is calculated for different curves, and the curve with maximum likelihood is selected for classification. Logistic regression determines the probability of the binary response.

    The line for linear regression is selected in such a way that the distance between the summation of all points and lines should be minimum. The equation for the plane is written as:
    \begin{equation}
    y=\omega^T x+b
    \end{equation}
    The trouble with linear regression is that we need to adjust the best fit line whenever new data comes. The exact coefficient is needed to find that will understand which is the best match line for that model, and then those particular coefficients or slopes will be adjusted to achieve the optimal plane. If x is the data point, then $\omega^T x$ is the distance between a particular data and plane, considering b=0.
    
    Distance between data point and the particular plane is :
        $ \sum\limits_{i=1}^n \omega^T x_i$
    
    Classes are classified according to the equation:
    \begin{equation}
    y_i\omega^T x_i \geq 0  \mathrm{\hspace{.5cm}correctly \hspace{.1cm} classified}
    \end{equation}
    
    \begin{equation}
     y_i\omega^T x_i \leq 0 \mathrm{\hspace{.5cm}incorrectly \hspace{.1cm} classified}
    \end{equation}
    
    To get the best fit line, cost function($ \sum\limits_{i=1}^n y_i \omega^T x_i$) should be maximum. Here, $\omega^T$ is the coefficient which needs to update  until maximum summation is found. The impact of an outlier has been avoided by adding the sigmoid function in the equation.Then the equation becomes:
    \begin{equation}
     max \sum \limits_{i=1}^n f(y_i\omega^T x_i)
    \end{equation}
     The sigmoid function is denoted as f in this equation.
     
In logistic regression, we tuned the parameter C, the regularization strength. For our proposed model, we used C = 1.

\subsubsection{Support vector machine(SVM)}
SVM or support vector machine is a supervised machine learning algorithm like linear regression, but here hyperplane with maximum marginal distance is selected to separate data into their categories.\cite{kibria2021comparative} SVM classifies positive and negative classes, and it has been shown in \ref{8} and \ref{9}:

    \begin{eqnarray}
      \mathbf x_i \mathbf w + b \geq +1,& y_i = +1
      \label{8} \\
      \mathbf x_i \mathbf w + b \leq -1,& y_i = -1
      \label{9} \\
      \equiv\\
      y_i(\mathbf x_i \mathbf w + b) - 1 \geq 0, & \forall i
    \end{eqnarray}
Here x and w are vectors, and bias is denoted by b. In SVM, it is important to optimize the marginal distance, which means the value of w  and b must be minimized to get the largest marginal distance  $||w||$. By optimizing the value of w and b, we can get our required distance to select the most fitted hyperplane.

    \begin{equation}
      min_w,_b,_\xi = \frac{1}{2}{\|w\|^2}
    \end{equation}
However, real data will have many overlaps, and if we follow the equation \ref{eqsvm}, then there will be over-fitting. Therefore a small amount of error is added to avoid over-fitting.       

   \begin{equation}
      min_w,_b,_\xi =  \frac{1}{2}{\|w\|^2} +C \displaystyle\sum^l_{i=1} \\ \xi_i
      \label{eqsvm}
   \end{equation}
Here, C is the number of errors that the model will consider, and $\xi$ is the value of error. A larger value of C indicates more error rate; if C is lower, there could be fewer errors \cite{burges1998tutorial}. In our model, C was 1 and 10 respectively for binary and multi-class.

\subsubsection{Decision tree(DT)}
A decision tree is a tree-like arrangement that is built based on attributes. The very top of the tree is called the root node or the root. Internal nodes are the nodes between the root and leaf node. Leaf nodes are the final class labels that represent the outcome. While constructing a decision tree, there is a concept or algorithm, which is ID3 algorithm. This algorithm tells that the first step for creating a decision tree is to select the right attributes for splitting the tree. To determine the best separation, we need to measure the impurity. A very popular model to calculate impurity is gini. Entropy is another model to estimate impurity by selecting which feature should be the root node. Choosing the right attribute for the root node will help to find the leaf node quickly and reduce time. A decision tree aims to get the leaf node quickly by selecting the right parameters. The equation for both gini and entropy is:
        \noindent 
        \begin{equation}
        Gini:Gini(E)=1-f(x)=1-\int_{j=1}^{n} Pj^2 
        \end{equation}
        \noindent 
        \begin{equation}
        Entropy:H(E)=1-f(x)=-\int_{j=1}^{n} Pj\log_2Pj
        \end{equation}
Information gain calculates the total entropy value from the top to the bottom node of the decision tree. The highest value of information gain is selected to construct the decision tree. Entropy is the calculation for one node, and information gain combines all the entropy. Entropy ranges between 0 to 1. 1 means the subset is completely impure where 0 is a pure subset. The construction of a DT is simple, and the implementation process is easy. The creation of a new node continues until a ground condition has not been formed.\cite{pei2019identification}.

Hyperparameter tuning is necessary for the decision tree. It is a value search method that will optimize the model. Our proposed model’s tuned parameters were maximum depth, the minimum number of sample leaves, and maximum features. This stopping criterion of these parameters helps to reduce over-fitting. In our decision tree model, maximum depth and maximum features were 8, the minimum sample leaf number was seven, and gini was used as the criterion.
                
\subsubsection{Random forest(RF)}
\begin{figure}[htbh]
    \centerline{\includegraphics[width=.6\columnwidth]{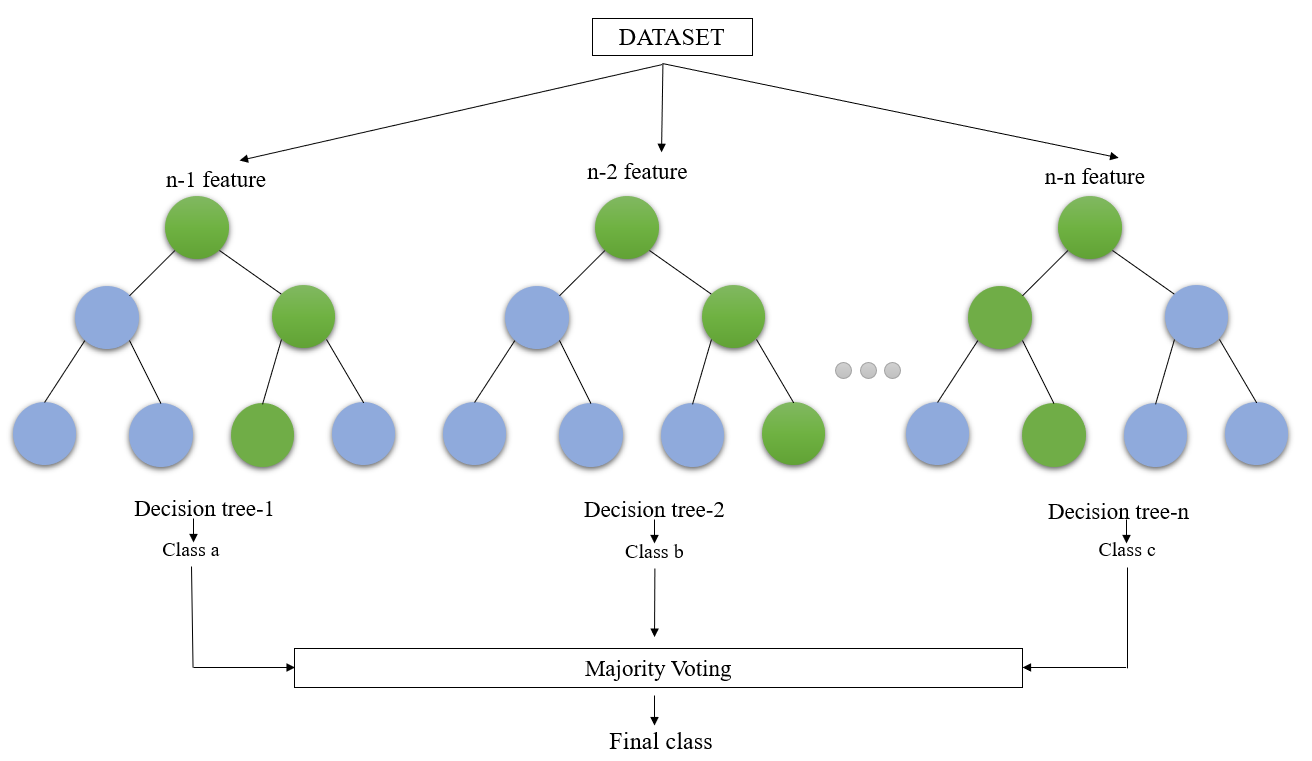}}
    \caption{Generalized structure for random forest}
    \label{rf}
    \end{figure}

The ensemble method combines multiple models to create an optimal model, and it is a machine learning technique. There are two forms of the ensemble model. Bagging and boosting. Random forest is a bagging technique. Bagging is also known as the bootstrap aggregating method. It has two main principles: row sampling and voting classifier. The given records are resampled and passed to the next base learner models to train. Aggregating is the voting classifier concept where, after training, output for test data will be selected for that class, which has the maximum vote from base learner models \cite{boulesteix2012overview}. Figure \ref{rf} shows a generalized model for the random forest.

A decision tree is based on a complete dataset, including all features and variables, where the random forest randomly chooses some row and unique features from the dataset to create several trees and then combines the result; the decision tree has two properties.

    \renewcommand\labelitemii{$\square$}
    \begin{itemize}
        \item Low bias: Since DT is built in its complete depth, training error is significantly less.
        \item High variance: The inability to classify test data leads to a larger amount of error.
    \end{itemize}
    
In a random forest, the high variance is converted into low variance due to the use of multiple decision trees. Adding new records in the dataset does not have a high impact on performance, so the accuracy does not decrease easily. In the regression problem, the random forest takes the mean or median of all decision trees, and in classifiers, it takes the majority vote for prediction. The number of estimators in our work was 100.

\subsubsection{Artificial neural network(ANN)} 

\begin{figure}[htbh]
\centering
\begin{subfigure}{.45\textwidth}
  \centering
  \includegraphics[width=.95\linewidth]{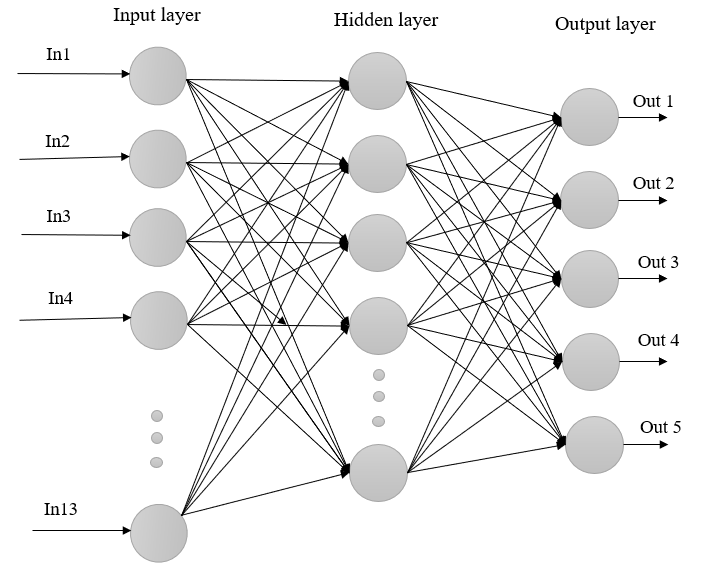}
  \caption{Multi-classification(five class) }
\end{subfigure}%
\begin{subfigure}{.45\textwidth}
  \centering
  \includegraphics[width=.95\linewidth]{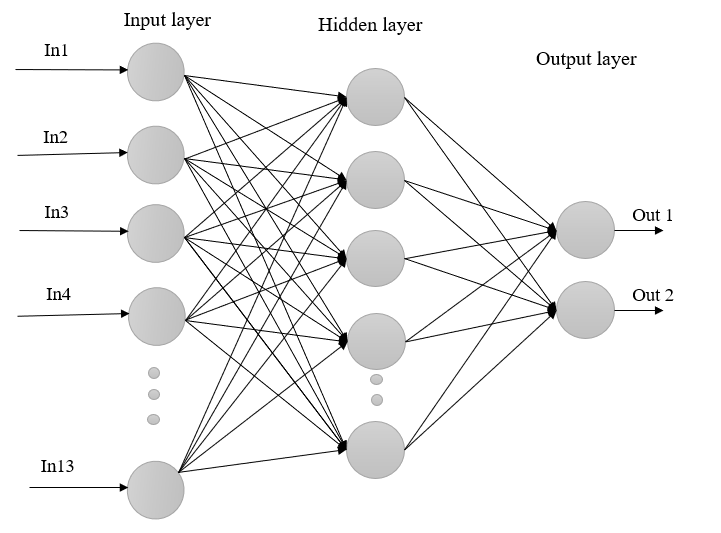}
  \caption{Binary classification }
\end{subfigure}
\caption{Architecture of artificial neural network  }
\label{ann}
\end{figure}

An artificial neural network is built on brain structure. Deep learning is a subfield of machine learning that uses algorithms inspired by the structure and function of brains. Such models used in deep learning are named artificial neural networks. So, artificial neural networks are computational structures that are inspired by the human brain. These neural networks are a collection of connected units called artificial neurons or only neurons. Each neuronal connection can transmit signals from one neuron to the next \cite{kibria2020efficient}. Typically neurons are organized in layers, and different layers perform different kinds of transformations depending on their inputs. Signals basically move from the first layer, called the input layer, to the last layer, called the output layer. Each layer between the input and the output layer is considered as a hidden layer. Figure \ref{ann} shows the structure for both binary and multi-classification of our model. The difference between them is that in multi-classification, the output layer contains five nodes, one for the absence of heart disease and the other four for reflecting the severity of the heart disease. On the other hand, the binary classification structure has two nodes for output.

Here, the dense layer was used to build the model. The first layer consisted of 13 neurons in our proposed model. Each of the nodes in the input layer reflects the individual features of each sample from the dataset. A weight has been allocated to each connection between the nodes. In training, the target tries to optimize the model's weights. The weights were continually updated to reach their optimal values. This optimization depends on the optimizer. We used stochastic gradient descent (SGD) in our model as the optimizer. SGD minimizes a given loss function. It assigns weights in such a way to make the loss function as close to zero as possible. The other hyper-parameters were batch size, learning rate, regularization. Batch size represents how many pieces of data we want to send to the model at once. The learning rate controls the amount of change in the model to calculate the error during the model weight update. Over-fitting occurs when the model is good at training samples but gives poor performance for testing. To prevent it, l2 regularization was used. Our artificial neural network contains one hidden layer for both binary and multi-classification. The number of neurons in the hidden layer was 8, and SGD was used as the optimizer.

\subsubsection{AdaBoost(ADA)}
In an adaptive boosting approach \cite{2yu2010identifying}, decision stump \cite{4shah2011feature} is used as a base learner. A stump is a tree with one node and two leaves, which is selected based on the entropy of the data given. The stump with the lowest entropy is chosen for this method. AdaBoost has three main characteristics. 

\renewcommand\labelitemii{$\square$}
\begin{itemize}
    \item To make a classification, AdaBoost combines a lot of weak learners (always stumps) to produce a strong algorithm. 
    \item Some stumps get more say in classification than others.
    \item Each stump is made by taking the previous stump’s mistakes into account.
\end{itemize}

The error made by the first base learner will be passed to the next base learner, and it will continue until the number of base learners that need to be created is declared. Boosting iterations are aimed at reducing the error of classification of the combined classification model over the training set. For classification, the maximum number of estimators was 200, and a .01 learning rate was used.

\subsubsection{Hyperparameter optimization}

We have optimized the six algorithms’ different parameters. The tuning of the parameters was different for both binary and multi-class classification. All optimal parameters have been listed in table \ref{optimal}.

\begin{table*}[h!]
\setlength{\tabcolsep}{3pt}
\centering
\caption{Optimal hyperparameters used in algorithms}
\begin{tabular}{|l|llll|}
\hline
\multicolumn{1}{|c|}{\multirow{3}{*}{\textbf{Algorithms}}} & \multicolumn{4}{c|}{\textbf{Optimal Parameters}}                                                                                                                                                                                                                                                                                                                                                                                                                                                                                                                        \\ \cline{2-5} 
\multicolumn{1}{|c|}{}                                     & \multicolumn{2}{c|}{Binary class}                                                                                                                                                                                                                                                                                                    & \multicolumn{2}{c|}{Multi class}                                                                                                                                                                                                 \\ \cline{2-5} 
\multicolumn{1}{|c|}{}                                     & \multicolumn{1}{c|}{\begin{tabular}[c]{@{}c@{}}train-test ratio\\  70:30\end{tabular}}                                                                          & \multicolumn{1}{c|}{\begin{tabular}[c]{@{}c@{}}train-test ratio \\ 80:20\end{tabular}}                                                                             & \multicolumn{1}{c|}{\begin{tabular}[c]{@{}c@{}}train-test ratio\\  70:30\end{tabular}}                                    & \multicolumn{1}{c|}{\begin{tabular}[c]{@{}c@{}}train-test ratio\\ 80:20\end{tabular}}                \\ \hline
\textbf{Logistic regression}                               & \multicolumn{1}{l|}{C=1}                                                                                                                                        & \multicolumn{1}{l|}{C=1}                                                                                                                                           & \multicolumn{1}{l|}{C=0.1}                                                                                                & C=0.001                                                                                              \\ \hline
\textbf{Support vector machine}                            & \multicolumn{1}{l|}{\begin{tabular}[c]{@{}l@{}}C=1\\ gamma=.1\end{tabular}}                                                                                     & \multicolumn{1}{l|}{\begin{tabular}[c]{@{}l@{}}C=1\\ gamma=.01\end{tabular}}                                                                                       & \multicolumn{1}{l|}{C=0.01}                                                                                               & C=100                                                                                                \\ \hline
\textbf{Decision tree}                                     & \multicolumn{1}{l|}{\begin{tabular}[c]{@{}l@{}}criterion=gini\\ maximum depth= 8\\ maximum features= 8\\ minimum sample leaf=7\\ splitter= random\end{tabular}} & \multicolumn{1}{l|}{\begin{tabular}[c]{@{}l@{}}criterion=gini \\ maximum depth= 8 \\ maximum features= 8\\ minimum sample leaf=7 \\ splitter= random\end{tabular}} & \multicolumn{1}{c|}{N/A}                                                                                                    & \multicolumn{1}{c|}{N/A}                                                                               \\ \hline
\textbf{Random forest}                                     & \multicolumn{1}{l|}{\begin{tabular}[c]{@{}l@{}}number of \\ estimators=100\end{tabular}}                                                                        & \multicolumn{1}{l|}{\begin{tabular}[c]{@{}l@{}}number of \\ estimators=100\end{tabular}}                                                                           & \multicolumn{1}{l|}{\begin{tabular}[c]{@{}l@{}}number of \\ estimators=100\end{tabular}}                                  & \begin{tabular}[c]{@{}l@{}}number of \\ estimators=200\end{tabular}                                  \\ \hline
\textbf{Artificial neural network}                         & \multicolumn{1}{l|}{\begin{tabular}[c]{@{}l@{}}learning rate=0.01 \\ optimizer=SGD \\ batch size=10\\ epoch=15\end{tabular}}                                    & \multicolumn{1}{l|}{\begin{tabular}[c]{@{}l@{}}learning rate=0.01\\ optimizer=SGD \\ batch size=10\\ epoch=15\end{tabular}}                                        & \multicolumn{1}{l|}{\begin{tabular}[c]{@{}l@{}}learning rate=0.01\\ optimizer=SGD\\ batch size=5\\ epoch=20\end{tabular}} & \begin{tabular}[c]{@{}l@{}}learning rate=0.01\\ optimizer=SGD\\ batch size=5\\ epoch=20\end{tabular} \\ \hline
\textbf{AdaBoost}                                          & \multicolumn{1}{l|}{\begin{tabular}[c]{@{}l@{}}number of \\ estimators=250 \\ learning rate =.01\end{tabular}}                                                  & \multicolumn{1}{l|}{\begin{tabular}[c]{@{}l@{}}number of \\ estimators=200\\ learning rate =.01\end{tabular}}                                                      & \multicolumn{1}{c|}{N/A}                                                                                                    & \multicolumn{1}{c|}{N/A}                                                                               \\ \hline
\end{tabular}
\label{optimal}
\end{table*}

\subsection{Performance Parameter}
Classification model performance is measured with the term of accuracy, precision, recall, and F1-score. The model's performance can be interpreted from the value of these parameters.
 Here, 
\begin{itemize}
     \item True positive ($T_P$): Outcome is positive, and prediction is positive.
    \item True negative ($T_N$): Outcome is negative, and prediction is negative.
    \item False positive ($F_P$): Outcome is negative, but prediction is positive. 
    \item False negative ($F_N$): Outcome is positive, but prediction is negative.
\end{itemize}

Accuracy is the measure of the percentage of correctly classified objects:
\begin{equation}
  Accuracy=\frac{T_P+T_N}{T_P+F_P+T_N+F_N}*100 
\end{equation}
Precision is also referred to as the false-positive rate. From precision, we get the number of correctly classified observations as positive to the total classified positive observations.
\begin{equation}
    Precision = \frac{T_P}{T_P+F_P} 
\end{equation}
Recall is often referred to as a truly positive rate. It is the ratio of total positive assumptions and the total amount of positive class attributes.
 \begin{equation}
    Recall = \frac{T_P}{T_P+F_N} 
\end{equation}

F1-score is the weighted average of precision and recall. When there is a need for the balance of precision and recall, it is required to take both the false positives and the false negatives into account. For an uneven class distribution, F1-score is generally more useful than accuracy.
\begin{equation}
  F1-score = \frac{2*Precision*Recall}{Precision+Recall} 
\end{equation}
ROC AUC score is also calculated as a classification parameter. It computes the area under the receiver operating characteristics curve using prediction scores.

To illustrate the classifiers’ performance, a confusion matrix has been used for the binary classification model. It summarizes the results of the predictions of a model. The matrix is given in table \ref{cm}.

\begin{table}[htbh]
\centering
\setlength{\tabcolsep}{3pt}
\caption{Confusion matrix}
\begin{tabular}{|c|c|c|}
\hline
                                                              & \begin{tabular}[c]{@{}c@{}}Actually\\  Positive\end{tabular}  & \begin{tabular}[c]{@{}c@{}}Actually\\  Negative\end{tabular}  \\ \hline
\begin{tabular}[c]{@{}c@{}}Predicted \\ Positive\end{tabular} & \begin{tabular}[c]{@{}c@{}}True Positive\\ $T_P$\end{tabular}  & \begin{tabular}[c]{@{}c@{}}False Positive\\ $F_P$\end{tabular} \\ \hline
\begin{tabular}[c]{@{}c@{}}Predicted\\  Negative\end{tabular} & \begin{tabular}[c]{@{}c@{}}False Negative\\ $F_N$\end{tabular} & \begin{tabular}[c]{@{}c@{}}True Negative\\ $T_N$\end{tabular}  \\ \hline
\end{tabular}
\label{cm}
\end{table}

\section{Result and Discussion}
Six algorithms were applied to the dataset. A grid search method was carried out to find the best-fit parameters for the algorithms. After training with those algorithms, three new fused models were developed using those models’ decision scores. A notable improvement was found in the result of all fused models. The key reason behind the fused model’s improvement is that if there is a miss classification in one algorithm, there is a  probability that the other algorithm may classify that particular data correctly. So after merging, there is a chance that we get the correct result for that specific case. This concept assists in giving the fused model a better efficiency.

The macro average was taken for the value of precision, recall, and F1-score in binary classification, and for multi-classification, the value of the weighted average was taken.

\begin{table}[htbh]
\centering
\setlength{\tabcolsep}{3pt}
\caption{Fusion Model-1 with logistic regression and random forest for multi classification}
\begin{tabular}{|c|c|c|c|c|c|}
\hline
\textbf{Ratio}         & \textbf{Algorithm} & \textbf{Acc}   & \textbf{Precision} & \textbf{Recall} & \textbf{F1-score} \\ \hline
\multirow{3}{*}{70:30} & LR                 & 59.34          & 63                 & 59              & 61                \\ \cline{2-6} 
                       & RF                 & 59.34          & 54                 & 59              & 56                \\ \cline{2-6} 
                       & \textbf{LR+RF}     & \textbf{65.93} & \textbf{66}        & \textbf{66}     & \textbf{66}       \\ \hline
\multirow{3}{*}{80:20} & LR                 & 72.13          & 64                 & 72              & 67                \\ \cline{2-6} 
                       & RF                 & 65.57          & 62                 & 66              & 63                \\ \cline{2-6} 
                       & \textbf{LR+RF}     & \textbf{75.41} & \textbf{72}        & \textbf{75}     & \textbf{71}       \\ \hline
\end{tabular}
\label{t3}
\end{table}

The result of multi-classification has been displayed in table \ref{t3},\ref{t4} and \ref{t5}. Table \ref{t3} shows the performance of LR, RF, and the combined model of them. After merging the two models’ decisions, the fused models’ performance parameters increased (LR+RF). It gave the highest accuracy of 75.41\% among the three fusion models.

\begin{figure}[htbh]
\centering
\begin{subfigure}{.33\textwidth}
  \centering
  \includegraphics[width=.92\linewidth]{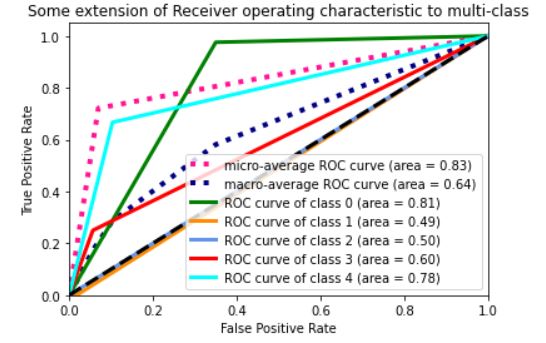}
  \caption{LR}
\end{subfigure}%
\begin{subfigure}{.33\textwidth}
  \centering
  \includegraphics[width=.92\linewidth]{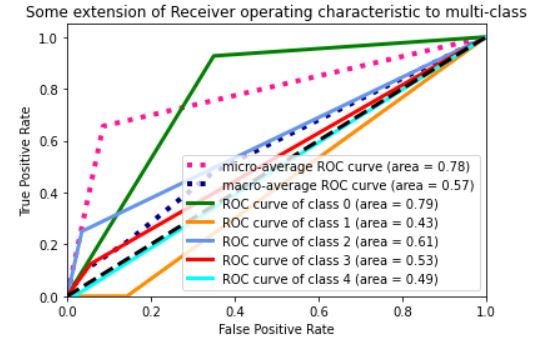}
  \caption{RF}
\end{subfigure}
\begin{subfigure}{.33\textwidth}
  \includegraphics[width=.92\linewidth]{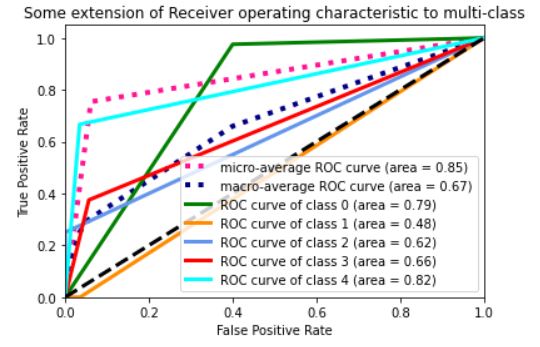}
  \caption{LR+RF}
\end{subfigure}
\caption{ROC curve of Model-1 for multi-classification(ratio 80:20)}
\label{roc7}
\end{figure}

Figure \ref{roc7} displays the Roc curve of logistic regression, random forest, and the fusion model(LR+RF) for five class classifications.

\begin{table}[htbh]
\centering
\setlength{\tabcolsep}{3pt}
\caption{Fusion Model-2  with support vector machine and artificial neural network for multi classification}
\begin{tabular}{|c|c|c|c|c|c|}
\hline
\textbf{Ratio}         & \textbf{Algorithm} & \textbf{Acc}   & \textbf{Precision} & \textbf{Recall} & \textbf{F1-score} \\ \hline
\multirow{3}{*}{70:30} & SVM                & 64.84          & 65                 & 65              & 63                \\ \cline{2-6} 
                       & ANN                & 65.93          & 70                 & 66              & 62                \\ \cline{2-6} 
                       & \textbf{SVM+ANN}   & \textbf{67.03} & \textbf{65}        & \textbf{67}     & \textbf{63}       \\ \hline
\multirow{3}{*}{80:20} & SVM                & 65.57          & 65                 & 66              & 64                \\ \cline{2-6} 
                       & ANN                & 68.85          & 72                 & 69              & 70                \\ \cline{2-6} 
                       & \textbf{SVM+ANN}   & \textbf{72.13} & \textbf{75}        & \textbf{72}     & \textbf{71}       \\ \hline
\end{tabular}
\label{t4}
\end{table}

Table \ref{t4} has represented the outcome of algorithm SVM and ANN. This fusion model provided 72.13\% accuracy, and the increment in the accuracy was 4\% and 7\% from ANN and SVM, respectively, at 80:20 split.

\begin{figure}[htbh]
\centering
\begin{subfigure}{.33\textwidth}
  \centering
  \includegraphics[width=.92\linewidth]{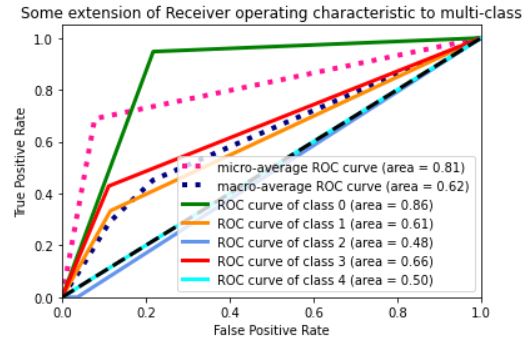}
  \caption{ANN}
\end{subfigure}%
\begin{subfigure}{.33\textwidth}
  \centering
  \includegraphics[width=.92\linewidth]{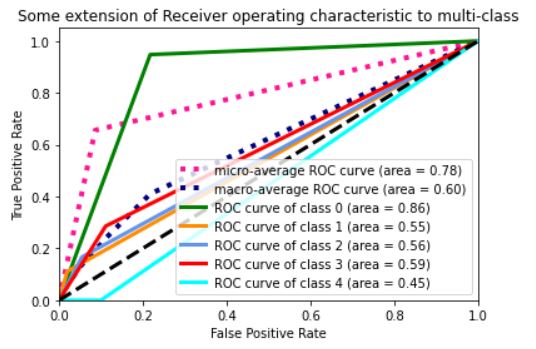}
  \caption{SVM}
\end{subfigure}
\begin{subfigure}{.33\textwidth}
  \includegraphics[width=.92\linewidth]{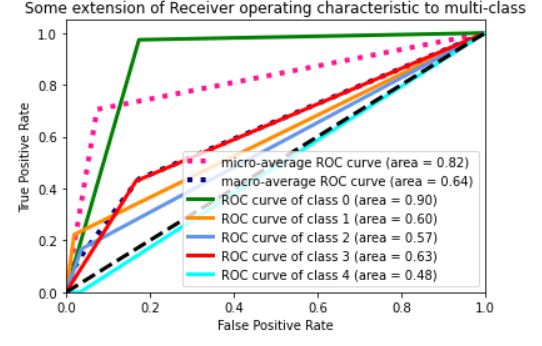}
  \caption{SVM+ANN}
\end{subfigure}
\caption{ROC curve of Model-2 for multi-classification(ratio 80:20)}
\label{roc8}
\end{figure}

The Roc curve of fusion model-2 and SVM and ANN curves have been displayed in figure \ref{roc8}.

\begin{table}[!h]
\centering
\setlength{\tabcolsep}{3pt}
\caption{Fusion Model-3  with artificial neural network and logistic regression for multi classification}
\begin{tabular}{|c|c|c|c|c|c|}
\hline
\textbf{Ratio}         & \textbf{Algorithm} & \textbf{Acc}   & \textbf{Precision} & \textbf{Recall} & \textbf{F1-score} \\ \hline
\multirow{3}{*}{70:30} & ANN                & 65.93          & 60                 & 66              & 62                \\ \cline{2-6} 
                       & LR                 & 62.64          & 62                 & 63              & 62                \\ \cline{2-6} 
                       & \textbf{ANN+LR}    & \textbf{67.03} & \textbf{62}        & \textbf{67}     & \textbf{64}       \\ \hline
\multirow{3}{*}{80:20} & ANN                & 70.49          & 70                 & 70              & 70                \\ \cline{2-6} 
                       & LR                 & 72.13          & 64                 & 72              & 67                \\ \cline{2-6} 
                       & \textbf{ANN+LR}    & \textbf{75.41} & \textbf{72}        & \textbf{75}     & \textbf{73}       \\ \hline
\end{tabular}
\label{t5}
\end{table}

\begin{figure}[!h]
\centering
\begin{subfigure}{.33\textwidth}
  \centering
  \includegraphics[width=.92\linewidth]{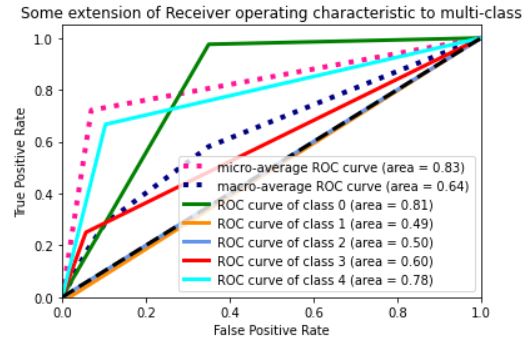}
  \caption{LR}
\end{subfigure}%
\begin{subfigure}{.33\textwidth}
  \centering
  \includegraphics[width=.92\linewidth]{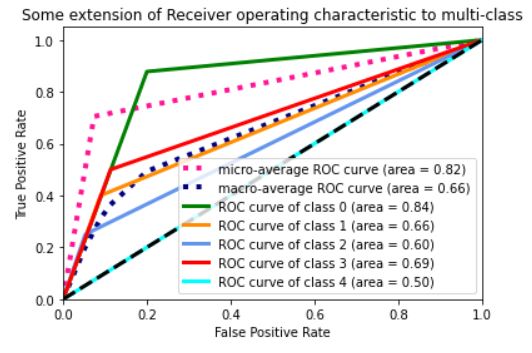}
  \caption{ANN}
\end{subfigure}
\begin{subfigure}{.33\textwidth}
  \centering
  \includegraphics[width=.92\linewidth]{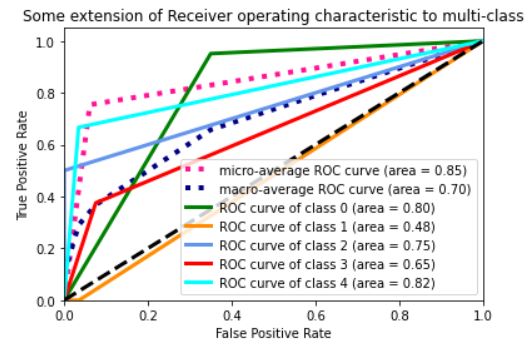}
  \caption{ANN+LR}
\end{subfigure}
\caption{ROC curve of Model-3 for multi-classification(ratio 70:30)}
\label{roc9}
\end{figure}

\begin{figure}[htbh]
\centering
\begin{subfigure}{.5\textwidth}
  \centering
  \includegraphics[width=.95\linewidth]{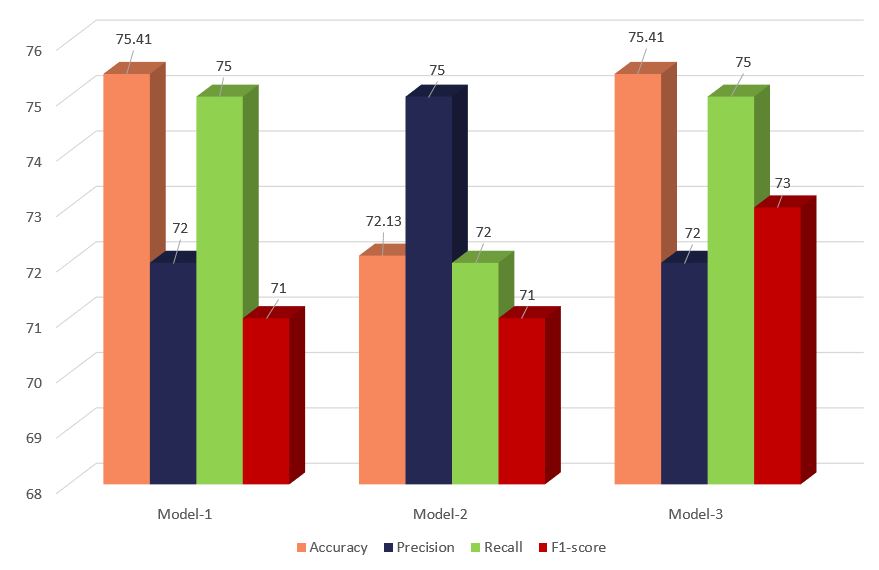}
  \caption{Multi-classification }
\end{subfigure}%
\begin{subfigure}{.5\textwidth}
  \centering
  \includegraphics[width=.95\linewidth]{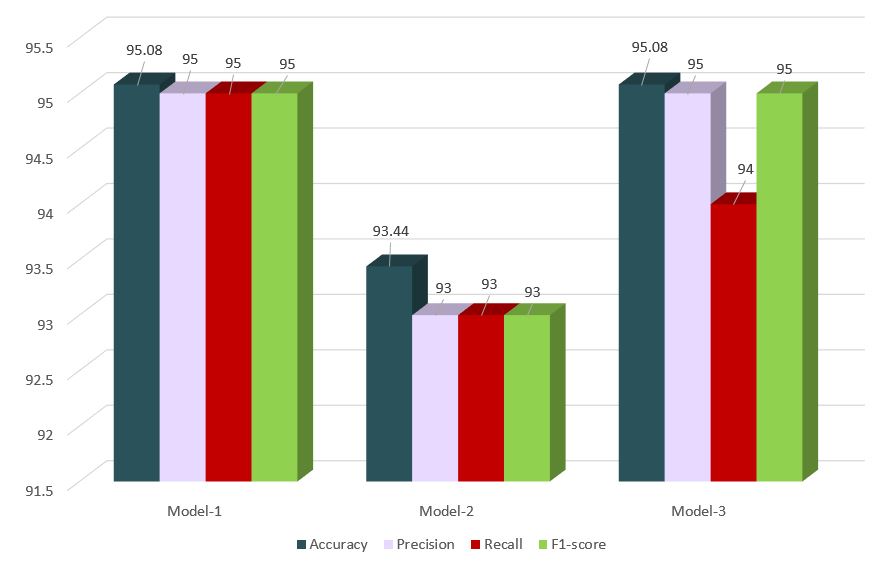}
  \caption{Binary classification}
\end{subfigure}
\caption{Comparison of weighted level fusion models (for 80:20 ratio) }
\label{compare}
\end{figure}

In table \ref{t5}, an accuracy of 75.41\% was obtained by fusion model-3, which was also the highest. But the precision did not improve for the 80:20 ratio of the hybrid model-3 (ANN+LR). As regards medical diagnosis, a false-negative is considered a criminal offense. This is why we put more focus on recall than on precision. The accuracy is 75.41\%. After merging, recall improved along with accuracy and f1-score. In figure \ref{roc9}, the roc curve of this model has been displayed. So these fusion systems provided a good performance overall. This comparison has been displayed in figure \ref{compare}. Four performance parameters were compared for the three multi-classification models in the figure.

\begin{table}[htbh]
\centering
\setlength{\tabcolsep}{3pt}
\caption{Fusion Model-1 with artificial neural network and random forest for binary classification}
\begin{tabular}{|c|c|c|c|c|c|c|c|c|}
\hline
\multirow{2}{*}{\textbf{Ratio}} & \multirow{2}{*}{\textbf{}}                                          & \textbf{Tp} & \textbf{Fp} & \multirow{2}{*}{\textbf{Acc}} & \multirow{2}{*}{\textbf{Prc}} & \multirow{2}{*}{\textbf{Recall}} & \multirow{2}{*}{\textbf{F1-score}} & \multirow{2}{*}{\textbf{\begin{tabular}[c]{@{}c@{}}Roc-Auc \\ score\end{tabular}}} \\ \cline{3-4}
                                &                                                                    & \textbf{Fn} & \textbf{Tn} &                                    &                                     &                                  &                                    &                                                                                    \\ \hline
\multirow{6}{*}{70:30}          & \multirow{2}{*}{ANN}                                               & 48          & 2           & \multirow{2}{*}{85.71}             & \multirow{2}{*}{87}                 & \multirow{2}{*}{86}              & \multirow{2}{*}{85}                & \multirow{2}{*}{84.59}                                                             \\ \cline{3-4}
                                &                                                                    & 11          & 30          &                                    &                                     &                                  &                                    &                                                                                    \\ \cline{2-9} 
                                & \multirow{2}{*}{RF}                                                & 47          & 3           & \multirow{2}{*}{91.21}             & \multirow{2}{*}{91}                 & \multirow{2}{*}{91}              & \multirow{2}{*}{91}                & \multirow{2}{*}{90.90}                                                             \\ \cline{3-4}
                                &                                                                    & 5           & 36          &                                    &                                     &                                  &                                    &                                                                                    \\ \cline{2-9} 
                                & \multirow{2}{*}{\begin{tabular}[c]{@{}c@{}}ANN\\ +RF\end{tabular}} & \textbf{48} & \textbf{2}  & \multirow{2}{*}{\textbf{93.41}}       & \multirow{2}{*}{\textbf{93}}        & \multirow{2}{*}{\textbf{93}}     & \multirow{2}{*}{\textbf{93}}       & \multirow{2}{*}{\textbf{93.59}}                                                    \\ \cline{3-4}
                                &                                                                    & \textbf{4}  & \textbf{37} &                                    &                                     &                                  &                                    &                                                                                    \\ \hline
\multirow{6}{*}{80:20}          & \multirow{2}{*}{ANN}                                               & 35          & 2           & \multirow{2}{*}{91.80}             & \multirow{2}{*}{92}                 & \multirow{2}{*}{91}              & \multirow{2}{*}{91}                & \multirow{2}{*}{91.10}                                                             \\ \cline{3-4}
                                &                                                                    & 3           & 21          &                                    &                                     &                                  &                                    &                                                                                    \\ \cline{2-9} 
                                & \multirow{2}{*}{RF}                                                & 34          & 3           & \multirow{2}{*}{93.44}             & \multirow{2}{*}{93}                 & \multirow{2}{*}{94}              & \multirow{2}{*}{93}                & \multirow{2}{*}{93.86}                                                             \\ \cline{3-4}
                                &                                                                    & 1           & 23          &                                    &                                     &                                  &                                    &                                                                                    \\ \cline{2-9} 
                                & \multirow{2}{*}{\begin{tabular}[c]{@{}c@{}}ANN\\ +RF\end{tabular}} & \textbf{35} & \textbf{2}  & \multirow{2}{*}{\textbf{95.08}}    & \multirow{2}{*}{\textbf{95}}        & \multirow{2}{*}{\textbf{95}}     & \multirow{2}{*}{\textbf{95}}       & \multirow{2}{*}{\textbf{94.61}}                                                    \\ \cline{3-4}
                                &                                                                    & \textbf{1}  & \textbf{23} &                                    &                                     &                                  &                                    &                                                                                    \\ \hline
\end{tabular}
\label{t6}
\end{table}

\begin{figure}[htbh]
\centering
\begin{subfigure}{.45\textwidth}
  \centering
  \includegraphics[width=.95\linewidth]{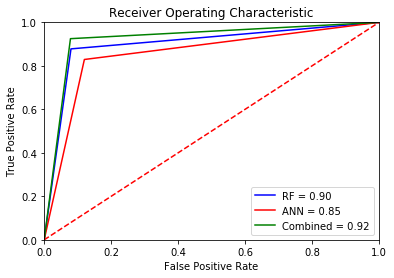}
  \caption{70:30 ratio }
\end{subfigure}%
\begin{subfigure}{.45\textwidth}
  \centering
  \includegraphics[width=.95\linewidth]{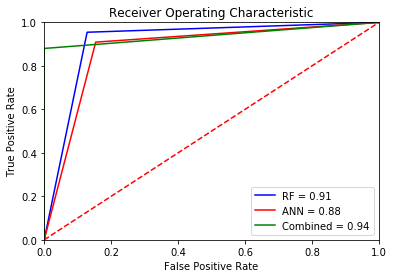}
  \caption{80:20 ratio}
\end{subfigure}
\caption{ROC curve of Model-1 for  binary classification  }
\label{roc10}
\end{figure}

Table \ref{t6},\ref{t7}, and \ref{t8} has the output of the binary classification. The performance parameters and the confusion matrix have been shown in these tables. Here we have gained the highest accuracy of 95.08\% from both fusion model-1(ANN+RF) and model-3(ADA+DT). Fusion model-1 has been displayed in table \ref{t6}.

\begin{table}[htbh]
\centering
\setlength{\tabcolsep}{3pt}
\caption{Fusion Model-2 with support vector machine and logistic regression for binary Classification}
\begin{tabular}{|c|c|c|c|c|c|c|c|c|}
\hline
\multirow{2}{*}{\textbf{Ratio}} & \multirow{2}{*}{\textbf{}}                                          & \textbf{Tp} & \textbf{Fp} & \multirow{2}{*}{\textbf{Acc}} & \multirow{2}{*}{\textbf{Prc}} & \multirow{2}{*}{\textbf{Recall}} & \multirow{2}{*}{\textbf{F1-score}} & \multirow{2}{*}{\textbf{\begin{tabular}[c]{@{}c@{}}Roc-Auc \\ score\end{tabular}}} \\ \cline{3-4}
                                &                                                                     & \textbf{Fn} & \textbf{Tn} &                                    &                                     &                                  &                                   &                                                                              \\ \hline
\multirow{6}{*}{70:30}          & \multirow{2}{*}{SVM}                                                & 50          & 4           & \multirow{2}{*}{87.91}             & \multirow{2}{*}{88}                 & \multirow{2}{*}{87}              & \multirow{2}{*}{87}               & \multirow{2}{*}{86.84}                                                       \\ \cline{3-4}
                                &                                                                     & 7           & 30          &                                    &                                     &                                  &                                   &                                                                              \\ \cline{2-9} 
                                & \multirow{2}{*}{LR}                                                 & 48          & 6           & \multirow{2}{*}{85.71}             & \multirow{2}{*}{85}                 & \multirow{2}{*}{85}              & \multirow{2}{*}{85}               & \multirow{2}{*}{84.98}                                                       \\ \cline{3-4}
                                &                                                                     & 7           & 30          &                                    &                                     &                                  &                                   &                                                                              \\ \cline{2-9} 
                                & \multirow{2}{*}{\begin{tabular}[c]{@{}c@{}}SVM\\ +LR\end{tabular}}  & \textbf{50} & \textbf{4}  & \multirow{2}{*}{\textbf{89.90}}    & \multirow{2}{*}{\textbf{89}}        & \multirow{2}{*}{\textbf{88}}     & \multirow{2}{*}{\textbf{89}}      & \multirow{2}{*}{\textbf{88.93}}                                              \\ \cline{3-4}
                                &                                                                     & \textbf{6}  & \textbf{31} &                                    &                                     &                                  &                                   &                                                                              \\ \hline
\multirow{6}{*}{80:20}          & \multirow{2}{*}{SVM}                                                & 34          & 1           & \multirow{2}{*}{91.80}             & \multirow{2}{*}{93}                 & \multirow{2}{*}{91}              & \multirow{2}{*}{91}               & \multirow{2}{*}{90.88}                                                       \\ \cline{3-4}
                                &                                                                     & 4           & 22          &                                    &                                     &                                  &                                   &                                                                              \\ \cline{2-9} 
                                & \multirow{2}{*}{LR}                                                 & 32          & 3           & \multirow{2}{*}{90.16}             & \multirow{2}{*}{90}                 & \multirow{2}{*}{90}              & \multirow{2}{*}{90}               & \multirow{2}{*}{89.95}                                                       \\ \cline{3-4}
                                &                                                                     & 3           & 23          &                                    &                                     &                                  &                                   &                                                                              \\ \cline{2-9} 
                                & \multirow{2}{*}{\begin{tabular}[c]{@{}c@{}}SVM\\  +LR\end{tabular}} & \textbf{33} & \textbf{2}  & \multirow{2}{*}{\textbf{93.44}}    & \multirow{2}{*}{\textbf{93}}        & \multirow{2}{*}{\textbf{93}}     & \multirow{2}{*}{\textbf{93}}      & \multirow{2}{*}{\textbf{93.30}}                                              \\ \cline{3-4}
                                &                                                                     & \textbf{2}  & \textbf{24} &                                    &                                     &                                  &                                   &                                                                              \\ \hline
\end{tabular}
\label{t7}
\end{table}

\begin{figure}[htbh]

\centering
\begin{subfigure}{.45\textwidth}
  \centering
  \includegraphics[width=.95\linewidth]{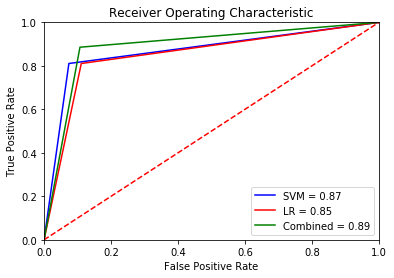}
  \caption{70:30 ratio }
\end{subfigure}%
\begin{subfigure}{.45\textwidth}
  \centering
  \includegraphics[width=.95\linewidth]{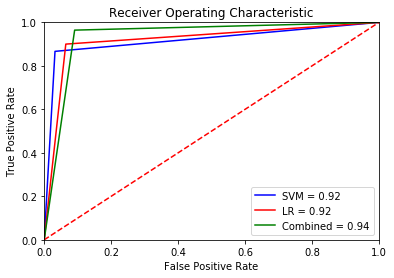}
  \caption{80:20 ratio}
\end{subfigure}
\caption{ROC curve of Model-2 for  binary classification  }
\label{roc11}
\end{figure}

Fusion model-2 and 3 also performed well. Table \ref{t7} represents model-2(SVM+LR), which gave 93.44\% accuracy with an 80:20 split ratio.

\begin{table}[htbh]
\centering
\setlength{\tabcolsep}{3pt}
\caption{Fusion Model-3 with adaboost and decision tree for binary classification}
\begin{tabular}{|c|c|c|c|c|c|c|c|c|}
\hline
\multirow{2}{*}{\textbf{Ratio}} & \multirow{2}{*}{\textbf{}}                                          & \textbf{Tp} & \textbf{Fp} & \multirow{2}{*}{\textbf{Acc}} & \multirow{2}{*}{\textbf{Prc}} & \multirow{2}{*}{\textbf{Recall}} & \multirow{2}{*}{\textbf{F1-score}} & \multirow{2}{*}{\textbf{\begin{tabular}[c]{@{}c@{}}Roc-Auc \\ score\end{tabular}}} \\ \cline{3-4}
                                &                                                                    & \textbf{Fn} & \textbf{Tn} &                                    &                                     &                                  &                                    &                                                                                    \\ \hline
\multirow{6}{*}{70:30}          & \multirow{2}{*}{ADA}                                               & 51          & 6           & \multirow{2}{*}{90.11}             & \multirow{2}{*}{89}                 & \multirow{2}{*}{90}              & \multirow{2}{*}{90}                & \multirow{2}{*}{90.32}                                                             \\ \cline{3-4}
                                &                                                                    & 3           & 31          &                                    &                                     &                                  &                                    &                                                                                    \\ \cline{2-9} 
                                & \multirow{2}{*}{DT}                                                & 56          & 1           & \multirow{2}{*}{90.11}             & \multirow{2}{*}{92}                 & \multirow{2}{*}{87}              & \multirow{2}{*}{89}                & \multirow{2}{*}{87.36}                                                             \\ \cline{3-4}
                                &                                                                    & 8           & 26          &                                    &                                     &                                  &                                    &                                                                                    \\ \cline{2-9} 
                                & \multirow{2}{*}{\begin{tabular}[c]{@{}c@{}}ADA\\ +DT\end{tabular}} & \textbf{53} & \textbf{4}  & \multirow{2}{*}{\textbf{92.31}}    & \multirow{2}{*}{\textbf{92}}        & \multirow{2}{*}{\textbf{92}}     & \multirow{2}{*}{\textbf{92}}       & \multirow{2}{*}{\textbf{91.61}}                                                    \\ \cline{3-4}
                                &                                                                    & \textbf{3}  & \textbf{31} &                                    &                                     &                                  &                                    &                                                                                    \\ \hline
\multirow{6}{*}{80:20}          & \multirow{2}{*}{ADA}                                               & 36          & 1           & \multirow{2}{*}{90.16}             & \multirow{2}{*}{91}                 & \multirow{2}{*}{88}              & \multirow{2}{*}{89}                & \multirow{2}{*}{88.23}                                                             \\ \cline{3-4}
                                &                                                                    & 5           & 19          &                                    &                                     &                                  &                                    &                                                                                    \\ \cline{2-9} 
                                & \multirow{2}{*}{DT}                                                & 33          & 4           & \multirow{2}{*}{90.16}             & \multirow{2}{*}{89}                 & \multirow{2}{*}{90}              & \multirow{2}{*}{90}                & \multirow{2}{*}{90.43}                                                             \\ \cline{3-4}
                                &                                                                    & 2           & 22          &                                    &                                     &                                  &                                    &                                                                                    \\ \cline{2-9} 
                                & \multirow{2}{*}{\begin{tabular}[c]{@{}c@{}}ADA\\ +DT\end{tabular}} & \textbf{36} & \textbf{1}  & \multirow{2}{*}{\textbf{95.08}}    & \multirow{2}{*}{\textbf{95}}        & \multirow{2}{*}{\textbf{94}}     & \multirow{2}{*}{\textbf{95}}       & \multirow{2}{*}{\textbf{95.19}}                                                    \\ \cline{3-4}
                                &                                                                    & \textbf{2}  & \textbf{22} &                                    &                                     &                                  &                                    &                                                                                    \\ \hline
\end{tabular}
\label{t8}
\end{table}

\begin{figure}[!h]
\vspace{7mm}
\centering
\begin{subfigure}{.45\textwidth}
  \centering
  \includegraphics[width=.95\linewidth]{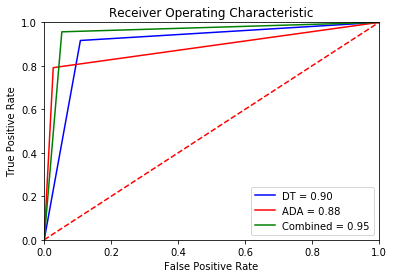}
  \caption{70:30 ratio }
\end{subfigure}%
\begin{subfigure}{.45\textwidth}
  \centering
  \includegraphics[width=.95\linewidth]{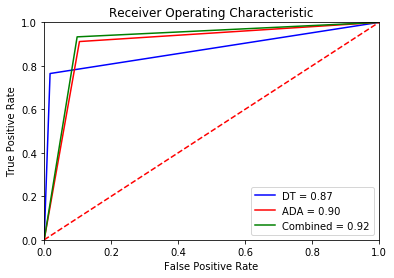}
  \caption{80:20 ratio}
\end{subfigure}
\caption{ROC curve of Model-3 for  binary classification  }
\label{roc12}
\end{figure}

The performance parameters of model-3 have been shown in table \ref{t8}. Here, model-3 provided an accuracy of 92.31\% and 95.08\% after fusion with the 70:30 and 80:20 split ratio, respectively. After fusion, the accuracy showed improvement compared to ADA and DT. 

In figure \ref{roc10},\ref{roc11}, \ref{roc12}, the roc curve for the binary fusion model-1,2 and 3 have been shown, respectively. 

Figure \ref{compare} shows the comparison chart for the three multi and binary classification models. All the models’ efficiency has significantly improved after fusion.

In table \ref{t9}, various approaches have been discussed along with their highest accuracy for diagnosing different diseases. Some of those works used the same dataset we used, and by analyzing them, we can say that our weighted fusion model’s performance is quite good comparing those. The research with two-class classification has done more than five classes, so we have discussed the previous work of binary classification of heart disease more than five class classifications.

\begin{table}[htbh]
\centering
\setlength{\tabcolsep}{3pt}
\caption{Comparison of various approaches for the diagnosis of  diseases}
\begin{tabular}{|l|l|l|}
\hline
Method                                                                                                                    & Results                                                                                                                                            & Ref      \\ \hline
\begin{tabular}[c]{@{}l@{}}Random forest, Decision tree, improved SVM\\ kernel\end{tabular}                               & \begin{tabular}[c]{@{}l@{}}The best result was achieved from improved SVM that  \\was 89.9\%; RF gave 82\% with an 80:20 split ratio.\end{tabular}           & \cite{harimoorthy2020multi} \\ \hline
\begin{tabular}[c]{@{}l@{}}Naive Bayes, multi layer perceptron, bagging\\  multi layer, and other algorithms\end{tabular} & \begin{tabular}[c]{@{}l@{}}The highest accuracy  obtained by Naive Bayes \\ was 92\% using 55 attributes\end{tabular}                               & \cite{hogo2020proposed}  \\ \hline
Deep learning, decision tree and others                                                                                   & \begin{tabular}[c]{@{}l@{}}Deep learning performed best among others with an  \\ accuracy of 98.07\% for diabetes prediction\end{tabular}          & \cite{naz2020deep}  \\ \hline
Logistic regression, random forest and others                                                                             & It gave 83\% accuracy using all features                                                                                                           & \cite{david2020machine}   \\ \hline
Support vector machine, logistic regression                                                                               & \begin{tabular}[c]{@{}l@{}}SVM \& LR gave 85\% \& 87\% accuracy respectively \\ with Cleveland dataset  (using 80\% data as training)\end{tabular}   &  \cite{Ramya2020}   \\ \hline
\begin{tabular}[c]{@{}l@{}}Support vector machine, multilayer perceptron \\ neural network\end{tabular}                   & \begin{tabular}[c]{@{}l@{}}Accuracy for multi classification was 68.86\% from\\ SVM using 80\% data for training (Cleveland dataset )\end{tabular} & \cite{nahiduzzaman2019prediction}    \\ \hline
Weighted fuzzy diagnostic system                                                                                          & \begin{tabular}[c]{@{}l@{}}This method gave 92.31\% accuracy for Cleveland’s  \\heart disease dataset using 70:30 train test ratio\end{tabular}    &  \cite{paul2018adaptive} \\ \hline
Hybrid method (ANN Fuzzy AHP)                                                                                             & Provided 91.10\% accuracy                                                                                                                          & \cite{samuel2017integrated} \\ \hline
\end{tabular}
\label{t9}
\end{table}

\section{Conclusion}
A significant percentage of the world's population is struggling with heart disease. To identify patients with heart disease, machine learning is necessary. This paper has focused on implementing a fused prediction model to identify patients with and without heart disease using machine learning algorithms. Binary classification classifies only the presence and absence of heart disease. In multi-classification, if heart disease is present, the severity can also be predicted. Our proposed models did both binary and multi-class classifications of heart disease. Multiple hybrid classification models have been designed to diagnose the disease, and then the distinction was made among them. The fusion models reduced the risk factors and improved the output in terms of accuracy, efficiency, and other factors. We saw a good improvement in the fusion model in both classifications, which makes the model more reliable. The decision from each algorithm was taken, and using the weighted sum rule, the scores were merged to form a new score. This new score is the decision score of our fusion model. Based on this score, the output will be predicted. The accuracy obtained for multi-classification from the fusion model is between 72-75\%, and for binary classification, the result is 93-95\%. After fusion, the performance improvement compared to the individual algorithm was about 1-3\%. This increase happened to all the parameters of the evaluation matrix.

A future real-world application of such work can be a mobile application, a web-based platform for physicians to update patient information. Patients would know the result with its risk factors easily. Depending on the severity, patients suffering from illness could be treated. This weighted fusion approach could also apply to other diseases as well.

\section{Code and Data Availability}
The Github link for the code and data are available  \href{https://github.com/hafsa-kibria/Weighted_score_fusion_model_heart_disease_prediction}{here}.

\bibliographystyle{unsrt}  
\bibliography{references}

\end{document}